\documentclass[runningheads]{llncs}
\usepackage[T1]{fontenc}
\usepackage{booktabs}
\usepackage{subcaption}
\usepackage{verbatim}
\usepackage{color}
\usepackage{amsfonts}
\usepackage{tikz}
\usepackage{caption}
\usetikzlibrary{arrows.meta, positioning, shapes, shadows.blur, fit, backgrounds}
\usepackage{graphicx} 
\usepackage{bbding}
\begin{document}
\title{\textit{KANEx}: Translating Kolmogorov-Arnold Networks' Interpretability to Medical Explainability}
\titlerunning{KANEx: Leveraging KANs for Medical Explainability}
%
\author{
Krithi Shailya\Envelope \inst{1} \and
Ananya Lakshmi Ravi\inst{1} \and
Venkatanathan K. V.\inst{1} \and
Sowmya S. Sundaram\inst{1} \and
Gokul S. Krishnan\inst{1} \and
Aditi Anand\inst{1,2}\thanks{Work done while the author was at the Centre for Responsible AI, IIT Madras} \and
Balaraman Ravindran\inst{1}
}

\authorrunning{Shailya et al.}
\institute{
Centre for Responsible AI, Wadhwani School of Data Science and AI,
Indian Institute of Technology Madras, Chennai, India\\
\email{krithishailya01@gmail.com}
\and
Vanderbilt University School of Medicine, Nashville, USA
}

  
\maketitle              
\begin{abstract}

Computer vision models have become highly effective for medical applications, yet their black-box nature continues to undermine clinician trust. In clinical workflows, chest X-ray classifiers are increasingly paired with Vision-Language Models (VLMs) to generate natural-language explanations. However, these systems add linguistic fluency without addressing the underlying opacity of the visual model. With the emergence of Kolmogorov-Arnold Networks (KANs), whose spline-based components provide inherently interpretable functional units, we investigate whether this architectural transparency can be leveraged to produce more trustworthy textual explanations. We introduce \textit{KANEx}, the first ever framework that leverages the symbolic transparency of KANs to ground VLM reasoning. This interpretability also made it possible to design \textit{KAN-Map}, a novel heatmap generation method derived directly from KAN models rather than gradient approximations.  We feed these grounded contexts into downstream VLMs for enhanced explainability. Benchmarked on the MIMIC-CXR dataset, we demonstrate that KAN-based architectures with ResNet/ViT baselines demonstrate improved semantic similarity while producing significantly more faithful saliency maps. KAN architectures improve visual localization and downstream reasoning quality by $\sim$10\%. Our findings suggest that grounding linguistic explanations and visual attributions in mathematically interpretable units is a necessary step toward trustworthy medical AI.


\end{abstract}

\section{Introduction \& Background}

Artificial Intelligence (AI) models, particularly computer vision models, are being deployed to assist clinicians, helping with tasks such as triage, prioritization, and diagnosis in medical settings \cite{intro}. Although these systems often achieve strong performance on benchmark datasets \cite{manzari2023medvit,manzari2025medical}, their internal decision-making processes are typically inaccessible and erodes clinician trust and sustained deployment. As a result, explainability is a pivotal requirement for medical AI models where transparency, and more importantly accountability are essential \cite{Wong2025}.

In this context, there has been a rise in the research landscape on explainable AI models for medicine \cite{survey1}. Prominent techniques include CAM \cite{zhou2016learning}, Grad-CAM \cite{selvaraju2017grad}, and occlusion sensitivity \cite{zeiler_visualizing_2013}, which are widely used to verify attention to clinically relevant pathology \cite{shi_survey_2024,adebayo_sanity_2020}. Vision-Language Models (VLMs) extend this approach by providing textual explanations, but they remain prone to hallucination and opaque reasoning \cite{Ghassemi2021}.

To improve transparency, we leverage interpretable Kolmogorov–Arnold Networks (KANs) \cite{liu2025kankolmogorovarnoldnetworks}, whose spline-based functions  reveal learned nonlinear mechanisms \cite{10.1145/3728637}. We introduce a novel approach, \textbf{KAN-Map}, a heatmap derived from KAN activations rather than gradients. Unlike Grad-CAM, which relies on a linear approximation of feature importance through gradient backpropagation, KAN-Map directly analyzes forward activations of spline functions to assess the contribution of each spatial patch. This approach eliminates the need for backward passes, yielding higher computational efficiency and capturing higher-order importance cues inherent in KAN’s nonlinear representations. Applied to chest X-ray classification, KAN-Map produces functionally grounded and faithful spatial attributions that better guide VLMs toward clinically consistent explanations. Our work presents the first concerted effort, to the best of our knowledge, to enhance the explainability of radiology reports on two fronts: (a) improved visual explainability using KAN-Map and (b) more robust explanations using VLMs that process these enhanced heatmaps.  

KAN-based variants improve localization IoU by $\sim$10--15\%, and improve semantic explanation quality (LExT~\cite{10.1145/3715275.3732104}) by up to $\sim$20\% over ResNet/ViT baselines. Our \textit{KAN-Map} further yields $\sim$25\% higher IoU, $>$20\% faithfulness gains, and $\sim$23\% LExT improvement over gradient-based methods.

\noindent Our contributions are threefold:

\begin{itemize}
    \item \textbf{KANEx}: A practical X-ray explainability pipeline which provides \textit{visual} heatmap localization, and \textit{textual} explanations in a single system. 
    \item \textbf{KAN-Map} A novel heatmap generation method built on the interpretable KAN components. 
    \item A first of a kind systematic empirical comparison of KAN variants of ResNet / ViT hybrid backbones within KANEx.
\end{itemize}





\section{Method}

We formalize \textit{multi-label chest X-ray diagnosis} with unified explanations and introduce \textbf{\textit{KANEx}}, a pipeline that realizes this framework. 

Let $x \in \mathbb{R}^{C \times H \times W}$, where $C$ is the number of channels, $H \times W$ the spatial dimensions, be an input chest X-ray image, $f(x)$ be a vision model that produces both classification and localization, $y \in \{0,1\}^K$ be a multi-label vector over $K$ clinical findings (e.g., pneumonia, effusion, cardiomegaly etc.)

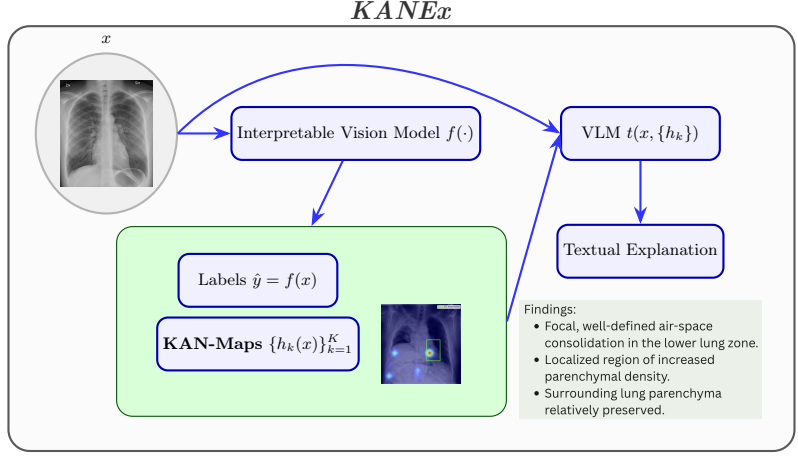
\begin{figure}[h!]
\centering
\begin{tikzpicture}[
    scale=0.7, transform shape,
    node distance=1.2cm and 1cm,
    every node/.style={font=\small},
    box/.style={
        rectangle, draw=blue!70!black, thick, rounded corners=4pt,
        minimum height=1cm, minimum width=3cm,
        fill=blue!6
    },
    imgbox/.style={
        ellipse, draw=gray!60, thick,
        minimum height=1.8cm, minimum width=1.8cm,
        fill=gray!12
    },
    arrow/.style={-Stealth, thick, blue!80},
    curvedarrow/.style={-Stealth, thick, blue!80, looseness=1.2}
]

\node[imgbox, inner sep=2pt] (img) {\includegraphics[height=2cm]{chest_xray_1.jpeg}};
\node[above=2pt of img] {$x$};

\node[box, right=of img] (kan)
    {Interpretable Vision Model $f(\cdot)$};

\node[box, below=1.75cm of kan, xshift=-1.85cm] (yhat)
    {Labels $\hat{y} = f(x)$};

\node[box, below=5pt of yhat] (hk)
    {\textbf{KAN-Maps} $\{h_k(x)\}_{k=1}^K$};

\node[right=0.3cm of hk] (heatmapimg) {\includegraphics[height=1.5cm]{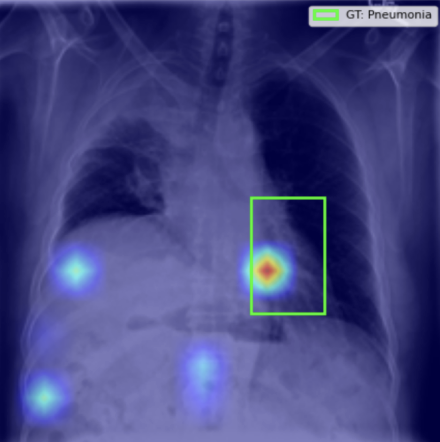}};

\node[box, right=1.5cm of kan] (vlm)
    {VLM $t(x,\{h_k\})$};



\node[box, below=of vlm] (out)
    {Textual Explanation};

\node[below=0.3cm of out] (expimg) {\includegraphics[scale=0.25]{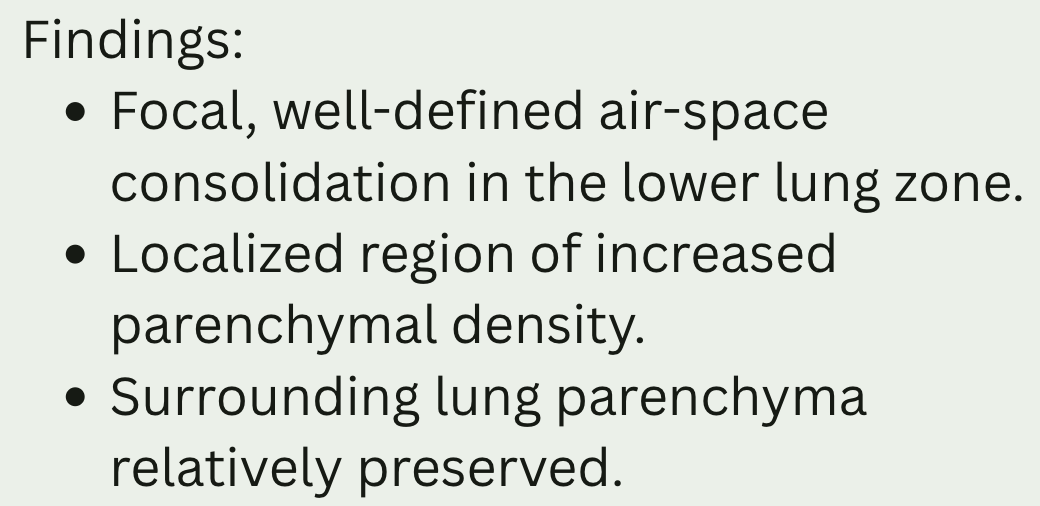}};

\begin{scope}[on background layer]
    \node[fit=(img)(kan)(yhat)(hk)(vlm)(out)(heatmapimg)(expimg),
         draw=gray!70!black, thick,
         fill=gray!3,
         rounded corners=8pt,
         inner sep=10pt,
         label={[black!80, font=\Large\bfseries]above:\textbf{\textit{KANEx} }}] (kanex) {};
    \node[
    draw=green!30!black,
    fill=green!15,
    rounded corners=6pt,
    inner sep=10pt,
    inner xsep=15pt,
    minimum width=4cm,
    minimum height=2cm,
    fit=(yhat)(hk)(heatmapimg)
    ] (group) {};
\end{scope}

\draw[arrow] (img) -- (kan);
\draw[arrow] (kan) -- (group.north);
\draw[arrow] (group.east) -- (vlm.west);
\draw[arrow] (vlm) -- (out);
\draw[arrow] (img.east) to[out=45,in=150] (vlm.west); 

\end{tikzpicture}

\caption{
\textit{Overview of the proposed pipeline KANEx}: An input image $x$ is processed by a KAN backbone $f(\cdot)$ to produce labels $\hat{y}$ and heatmaps $\{h_k(x)\}_{k=1}^K$. These outputs are provided to a VLM which generates textual explanations.
}
\label{fig:pipeline}
\end{figure}

Our goal is to develop  a pipeline (Figure~\ref{fig:pipeline}) that, for each image $x$, produces: (a) A set of predicted probabilities $f(x) = \hat{y} = (\hat{y}_1,\dots,\hat{y}_K)$ over the $K$ labels, (b) Spatial explanations in the form of novel heatmaps derived from KAN importance (\textbf{\textit{KAN-Map}}) $h_k(x)$ that localize image regions supporting each predicted label $k$. (c) A \textit{single, holistic textual explanation} $t(x, \{h_k\}_k)$ formed by attention-weighted aggregation $\sum_k \hat{y}_k \cdot h_k(x)$ or multi-channel VLM input, synthesizing evidence across all labels into one practitioner-friendly description. 

\subsection{Interpretable Models: Kolmogorov-Arnold Networks}

Standard vision models for chest X-ray analysis include ResNets  and vision transformers (ViTs) \cite{Xu2023,manzari2025medical}. ResNets \cite{he2016deep} use residual connections with convolutional filters to capture local textures and patterns. Vision transformers (ViTs)~\cite{dosovitskiy2021imageworth16x16words} treat images as sequences of patches, applying self-attention to capture long-range dependencies. We integrate the popular Kolmogorov--Arnold Networks (KANs) \cite{liu2025kankolmogorovarnoldnetworks} into these architectures, replacing neural layers with spline-based edge functions for providing \emph{intrinsic} interpretability through mathematically structured representations. Unlike MLPs that apply fixed nonlinearities to linear combinations at each node,
$x_i^{(l+1)} = \sigma \left( \sum_j w_{ij}^{(l)} x_j^{(l)} \right)$,
KANs replace learnable weights $w_{ij}$ with univariate learnable functions $\phi_{i,j}$ on network edges:
$x_i^{(l+1)} = \Phi_i \left( \sum_{j=1}^{n_l} \phi_{i,j}^{(l)} \left( x_j^{(l)} \right) \right)$, where $\Phi_i$ is a fixed nonlinearity and each $\phi_{i,j}$ is typically parameterized as a B-spline. This structure enables visualization of spline functions and symbolic simplification, exposing interpretable decision rules~\cite{somvanshi2025survey}. We explore several KAN variants integrated into ResNet and ViT architectures. \noindent We consider three KAN variants: \textbf{VanillaKAN}, which uses spline-based edge functions~\cite{liu2025kankolmogorovarnoldnetworks}; \textbf{GroupKAN}, which groups splines for improved high-dimensional efficiency~\cite{li2025groupkan}; and \textbf{RationalKAN}, which replaces splines with rational activation functions to enhance expressivity~\cite{aghaei2024rkan}. 

\subsection{KAN-Map: Spline-Derived Heatmaps}

Heatmaps provide visual explanations by highlighting regions that drive model predictions. Traditional Class Activation Mapping (CAM)~\cite{zhou2016learning} assumes linear classifiers, while Grad-CAM~\cite{selvaraju2017grad} relies on gradient approximations. We extend this idea to KAN classifiers by directly evaluating the learned spline functions over spatial feature vectors. We pass the visual features from each image region through the KAN's learned spline functions and measure how strongly that region supports the prediction, using these scores to generate the heatmap.

Let $\mathbf{F} \in \mathbb{R}^{C \times H \times W}$ denote the final convolutional feature map, where $C$ is the channel dimension and $H \times W$ are spatial dimensions. For each spatial location $(x,y)$, the feature vector $\mathbf{F}_{x,y} \in \mathbb{R}^C$ is passed through the trained KAN classifier. The class-$k$ heatmap value is obtained as
\[
h_k(x,y) = \phi_{2,k}\!\left(\phi_{1}(\mathbf{F}_{x,y})\right),
\]
where $\phi_1: \mathbb{R}^C \rightarrow \mathbb{R}^D$ and $\phi_2: \mathbb{R}^D \rightarrow \mathbb{R}^K$ denote the first and second KAN spline layers, respectively. Each univariate spline unit is parameterized using a B-spline basis with grid size $G{=}5$ and order $p{=}3$:
\[
\phi_{j,c}(z) = \sum_m w_{j,c,m}\, B_m(z),
\]
where $B_m$ are B-spline basis functions and $w_{j,c,m}$ are learned coefficients.

\textbf{Implementation:} The feature map is reshaped to $(HW)\!\times\!C$ and forward-passed through the trained KAN head to obtain class logits for every spatial location. The logits corresponding to class $k$ are reshaped to $H \times W$, passed through ReLU, and normalized to $[0,1]$ to obtain the final heatmap. This forward-only procedure requires no gradients or linear approximations and directly reflects the learned spline mappings.

\subsection{Prompting the Vision Language Model (VLM)}
The VLM receives saliency-enhanced images along with the predicted diagnosis as input and is prompted to generate a natural-language explanation justifying that diagnosis. We design a detailed prompt that encompasses the medical context, the X-ray, the heatmap and prompt radiology report generation.


\noindent
\fbox{
    \begin{minipage}{\linewidth}
    \footnotesize
        \textbf{Prompt for extracting explanations from normal/enhanced images:}
        
        \textbf{Input:} A chest X-ray image and the target diagnosis label.

\textbf{Instruction:} Analyze the provided chest X-ray and generate a structured radiology report using language appropriate for a physician. The diagnosis for this case is \texttt{\{diagnosis\}}.

\textbf{Output format:}

\textit{Findings:} Describe the radiographic abnormalities that support the given diagnosis, or state if findings are subtle.

\textit{Explanation:} Explain, using expert radiologic reasoning, how the imaging findings support the diagnosis and discuss relevant differential considerations if applicable.

\textbf{Example:}
<Example taken from ground truth>

Now generate a report for the given chest X-ray using the same format







    \end{minipage}
}
    

\section{Experiments \& Results}
We describe further details of the dataset, experimental setup and metrics below.


\paragraph{\textbf{Custom Vision-Language Dataset for Explainability:}}

We constructed a custom multimodal dataset to comprehensively evaluate visual explainability and explanation quality in our experiments. We use a subset of the chest radiographs from MIMIC-CXR \cite{PhysioNet-mimic-cxr-2.1.0,Johnson2019}. These were matched, via subject and study identifiers, to structured diagnosis codes from MIMIC-IV and narrative “Brief Hospital Course” notes from MIMIC-IV-Note. For each matched case, we extracted the X-ray and radiology findings, CheXpert labels for training, retrieved note passages referencing the X-ray, and synthesized these elements into a single clinically grounded natural language rationale. For evaluation, we used a subset of data points which were derived from MS-CXR \cite{PhysioNet-ms-cxr-1.1.0} that provided bounding boxes for MIMIC data for heatmap evaluation. This process resulted in a total of 20k cases, each with an image, a diagnosis, a segmentation, and a comprehensive ground-truth explanation. From these cases, 30\% were held out as an unseen test set for evaluation.

\paragraph{\textbf{Experimental Setup:}}


For the vision backbones, we used both ResNet and Vision Transformer (ViT) architectures as baselines. To study the interaction between KANs and existing architectures in a controlled manner, we replaced the final MLP classification head with the KAN variants while freezing the pretrained backbone in all experiments. This design choice ensured that any changes in explanation quality or interpretability could be attributed specifically to the KAN-based head rather than differences in feature extraction or overall model capacity.  All backbone weights, training configurations, and optimization settings were kept identical across models to maintain a fair and consistent comparison. To evaluate visual explainability, we compared our proposed \textbf{KAN-Map} with two other methods: Grad-CAM \cite{selvaraju2017grad}, gradient-weighted attention rollout (Attn-R)\cite{jo2025gmargradientdrivenmultiheadattention}.

For generating explanations, we used LLaVA (Large Language and Vision Assistant) \cite{liu_visual_2023}, an open-source instruction-tuned multimodal model built on the LLaMA backbone, due to its strong image–text reasoning ability to generate structured long-form explanations aligned with physician reports, and because prior evidence indicates that general-purpose models often outperform finetuned medical models in generative explanation quality \cite{10.1145/3715275.3732104}. The model was prompted to respond as a physician interpreting a chest X-ray for a clinical audience, ensuring consistent tone and structure with the ground truth. The curated dataset, prompts used, the modified model architectures and training configurations are available in our code base\footnote{{\url{https://github.com/cerai-iitm/KANEx}}}.

\paragraph{\textbf{Evaluation Metrics:}}



\begin{figure}[h]
  \centering
   \includegraphics[width=0.95\textwidth]{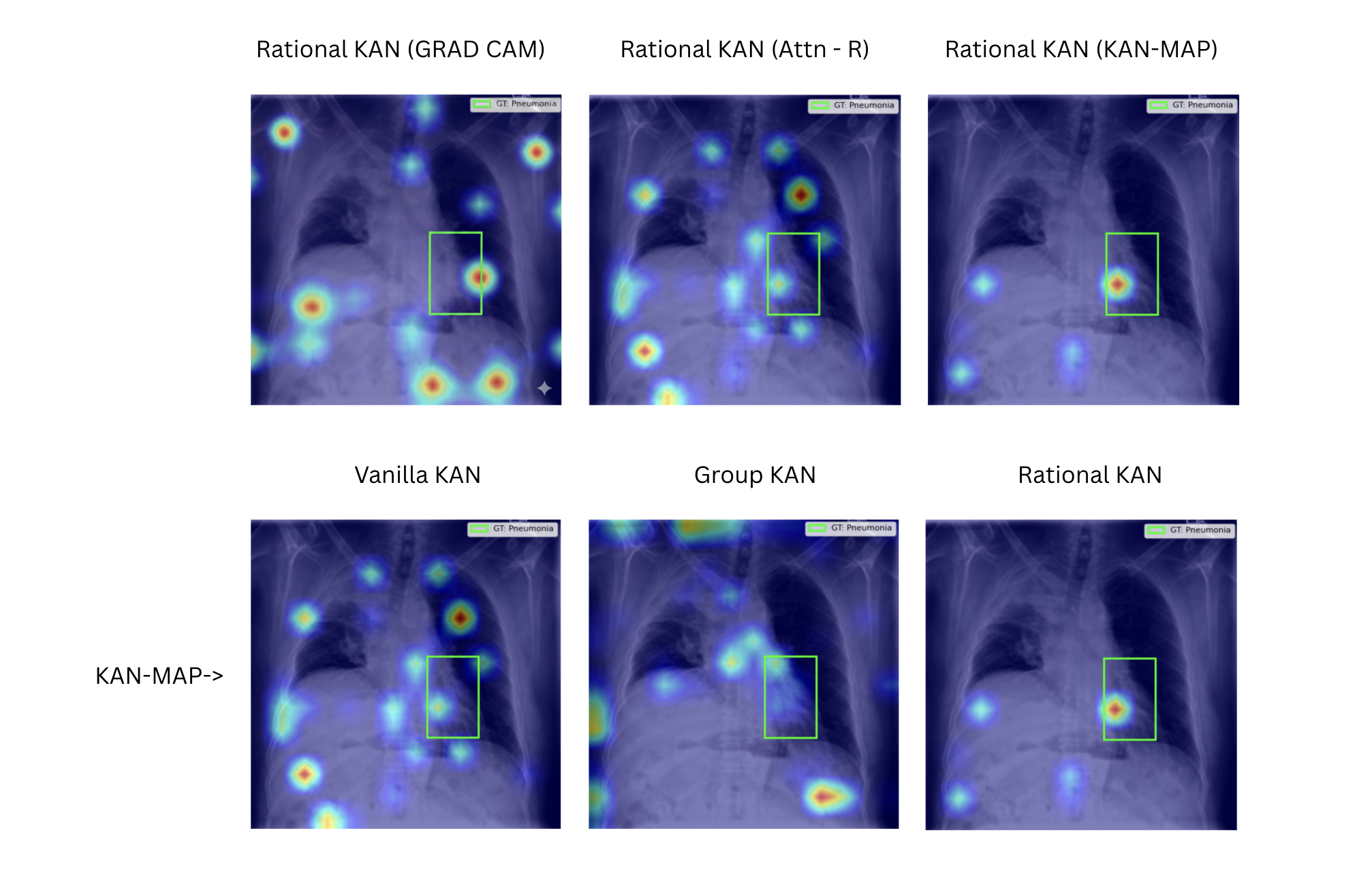}
    
  \caption{Ablation Studies for KAN-Map: Localization methods for Rational KAN (Grad-CAM, Attn-R, KAN-Map) with ground truth bounding boxes (top) and KAN-Map across VanillaKAN, GroupKAN, RationalKAN (bottom)}
  \label{fig:four-panels}
\end{figure}
\begin{figure}[h!]
  \centering
  \begin{subfigure}[b]{0.7\textwidth}
    \centering
    \includegraphics[width=\linewidth]{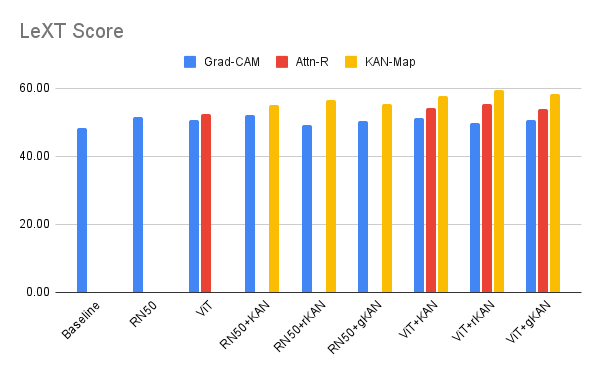}
    \label{fig:sub1}
  \end{subfigure}\hfill
  \begin{subfigure}[b]{0.3\textwidth}
    \centering
    \begin{subfigure}[t]{\linewidth}
      \centering
      \includegraphics[width=\linewidth]{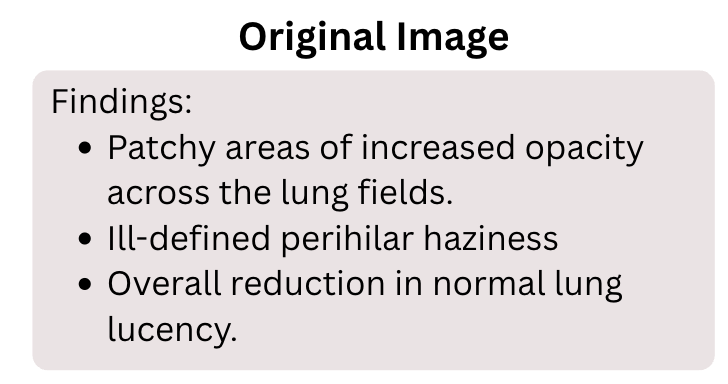}
      \label{fig:sub2}
    \end{subfigure}\par\vspace{0.2em}
    \begin{subfigure}[t]{\linewidth}
      \centering
      \includegraphics[width=\linewidth]{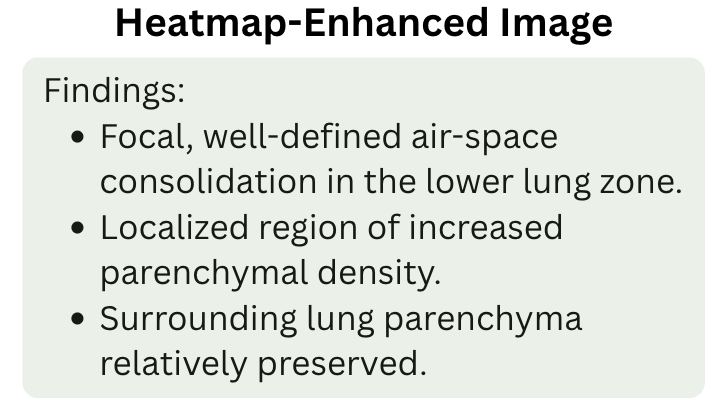}
      \label{fig:sub3}
    \end{subfigure}
  \end{subfigure}
  \caption{LExT bar plot comparing baseline, RN/ViT, and KAN variants; corresponding explanations reveal sub-par baseline explanation (top-right) vs. improved KAN-Map explanation (bottom-right).}
  \label{fig:two-panels}
\end{figure}

We evaluate model performance across two complementary dimensions (Table~\ref{tab:metrics}): \textit{Visual Explainability}  and \textit{Explanation Quality}. We examine localizations (Table~\ref{tab:results-iou}) using IoU, Energy@10, Area@50 and Faithfulness (Table~\ref{tab:metrics})  across Grad-CAM (G), and our KAN-Map (K) method. In the case of ViTs, we additionally compare with the Attn-R (A) method. We also present some of (Figure~\ref{fig:four-panels}) the heatmaps. Finally, we compare the text explanation quality generated by the VLM directly against the various KAN types with the three different heatmap configurations (Grad-CAM, Attn-R, KAN-Map) using LExT-C, which provides a NER-based embedding overlap to align  We also look at how the explanations differ when we use KAN-Map based images versus the baseline (Figure~\ref{fig:two-panels}). Friedman test was run to compare model distributions on IoU. The distributions are significantly different with $p<0.001$.

\begin{table}[h!]
\centering
\footnotesize
\caption{Definitions of evaluation metrics used in the proposed fraemwork}
\label{tab:metrics}
\renewcommand{\arraystretch}{1.15}
\begin{tabular}{p{2.1cm} p{2cm} p{8cm}}
\toprule
\textbf{Category} & \textbf{Metric} & \textbf{Definition (Range, Direction)} \\
\midrule

Visual \newline Explainability & IoU & Intersection-over-Union between predicted heatmaps and ground-truth bounding boxes  \\
 & Energy@10 & The fraction of total heatmap energy contained within the top 10\% most salient pixels \cite{selvaraju2017grad} \\
 & Area@50 &  The fraction of image area required to capture 50\% of total heatmap energy \cite{selvaraju2017grad}   \\
  & Faithfulness & We assess explanation faithfulness as the difference between \textit{Deletion AUC}, which measures the drop in model confidence when salient regions are progressively removed and \textit{Insertion AUC}, which measures confidence recovery when salient regions are gradually introduced \cite{selvaraju2017grad}. \\\hline
 Explanation Quality & LExT-C & The correctness subset of LExT measures clinical alignment and correctness between generated explanations and reports through lexical and factual overlap~\cite{10.1145/3715275.3732104}.  \\

\bottomrule
\end{tabular}
\end{table}


  



\begin{table}[h]
\centering
\caption{Visual explainability comparison across RN50, ViT, and their KAN variants. Bold values indicate the best performance within each backbone family and explanation method.}
\label{tab:results-iou}
\resizebox{\textwidth}{!}{
\begin{tabular}{l *{12}{c}}
\toprule
 & \multicolumn{3}{c}{\textbf{IoU $\uparrow$}} & 
   \multicolumn{3}{c}{\textbf{Area@50 $\downarrow$}} &
   \multicolumn{3}{c}{\textbf{Energy@10 $\uparrow$}}&
   \multicolumn{3}{c}{\textbf{Faithfulness $\uparrow$}} 
   \\
\midrule
\textbf{Model} & G & A & K 
& G & A & K 
& G & A & K 
& G & A & K \\
\midrule
RN50 
& 0.062 & -- & -- 
& 0.184 & -- & -- 
& 0.268 & -- & -- 
& 3.910 & -- & -- \\

ViT 
& 0.066 & 0.067 & -- 
& 0.011 & 0.016 & -- 
& 0.392 & 0.381 & -- 
& 1.050 & 2.450 & -- \\

\midrule
RN50(KAN)
& 0.060 & -- & 0.069 
& 0.166 & -- & 0.075 
& 0.280 & -- & 0.568 
& 3.893 & -- & 9.116 \\

RN50(rKAN)
& 0.061 & -- & \textbf{0.076} 
& 0.197 & -- & 0.114 
& 0.257 & -- & 0.432 
& \textbf{5.554} & -- & \textbf{9.789} \\

RN50(gKAN)
& \textbf{0.063} & -- & 0.075 
& \textbf{0.151} & -- & \textbf{0.075} 
& \textbf{0.340} & -- & \textbf{0.570 }
& 3.086 & -- & 9.300 \\

\midrule
ViT(KAN)
& 0.067 & 0.068 & 0.079 
& 0.010 &\textbf{ 0.017} & 0.023
& 0.388 & 0.376 & 0.365 
& 0.960 & 3.330 & 6.200 \\

ViT(rKAN) 
& 0.068 & 0.070 & \textbf{0.079}
& 0.010 & 0.019 & 0.025 
& 0.386 & 0.372 & 0.358 
& \textbf{1.010 }& \textbf{4.850 }& \textbf{7.500} \\

ViT(gKAN)
& \textbf{0.070} & \textbf{0.074} & 0.077 
& \textbf{0.008} & \textbf{0.017} &\textbf{ 0.021}
& \textbf{0.390} & \textbf{0.379} &\textbf{ 0.367} 
& 0.840 & 4.050 & 6.270 \\

\bottomrule
\end{tabular}}
\end{table}






\subsection{Discussion}

\textbf{KAN-Map} consistently yielded improved localization and faithfulness compared with gradient-based baselines. Its evaluation of learned spline mappings at the spatial level produced heatmaps that were both more compact and more aligned with class-relevant behavior, demonstrating the benefit of incorporating KAN-aware spatial structure into interpretability evaluation.

The \textit{Inter-KAN} analysis reveals distinctive performance patterns among the KAN variants. The GroupKAN model achieved the best overall classification and alignment with its internal function, attaining the highest KAN-Map Faithfulness and LExT scores. While RationalKAN reached the highest IoU with KAN-Map, the VanillaKAN and GroupKAN variants generated more compact and concentrated heatmaps, indicated by the lowest \textit{Area@50} and highest \textit{Energy@10} scores. This observation suggests a trade-off in interpretability: the most spatially accurate localization does not always correspond to the most focused explanatory pattern. Complementarily, it also helps translate theoretical interpretability: RationalKAN’s rational spline functions provide smoother and more globally stable function approximations that improve attribution faithfulness, while GroupKAN’s grouped basis decomposition preserves spatial feature separability, enabling sharper and more localized activations that translate into higher IoU and localization quality.

 We also observe that KAN-Map based localisations (Figure \ref{fig:two-panels}) generally produce higher-quality textual explanations than GRAD-CAM and KAN variants achieve more plausible alignment between generated explanations and model reasoning, while the absolute localization accurazy indicates potential for enhancing the spatial alighment of explanations. Several interpretative factors merit consideration when assessing these findings. The evaluation of faithfulness depends on the chosen perturbation and inpainting strategies, and alternate methodologies could lead to different quantitative outcomes.
Our work thus fits in with the growing literature on semantic alignment and explainability, especially in the field of radiology reports \cite{thawakar2024xraygpt,gu2025radalign,deperrois2025radvlmmultitaskconversationalvisionlanguage}.



To further underscore the trustworthiness and practical value of our pipeline, we plan to evaluate the proposed method with clinician oversight to assess its interpretability and relevance in real-world diagnostic contexts. Our findings encourage the adoption and further fine-tuning of KAN-based variants to more fully leverage their intrinsic functional interpretability, providing researchers, clinical-AI developers, and radiologists with a principled framework for explanation auditing, faithful VLM-grounded reporting, model debugging, and ultimately safer and more transparent clinical deployment.

\section{Conclusion}
In summary, we introduce a unified pipeline combining KANs with VLMs for multi-label chest X-ray diagnosis and practitioner-useful textual explanations. Evaluating VanillaKAN, GroupKAN, and RationalKAN variants across ResNet and ViT backbones, we show how intrinsic interpretability, exposed by our novel KAN-Map heatmaps, enables linking findings to localized image evidence. This work bridges model interpretability, visual grounding, and clinical utility, working towards providing a practical framework for trustworthy medical AI.

\begin{credits}

\subsubsection{\discintname}
The authors have no competing interests to declare that are relevant to the content of this article.

\end{credits}
\bibliographystyle{splncs04}
\bibliography{Paper-4330}








\end{document}